\documentclass[letterpaper, 10 pt, conference]{ieeeconf}  

\IEEEoverridecommandlockouts                              

\usepackage{graphics} 
\usepackage{epsfig} 
\usepackage{mathptmx} 
\usepackage{times} 
\usepackage{amsmath} 
\usepackage{amssymb}  
\usepackage{cite}
\usepackage{color}
\usepackage{amsfonts}

\title{\LARGE \bf
MVFusion: Multi-View 3D Object Detection with Semantic-aligned \\ Radar and Camera Fusion
}

\author{Zizhang Wu$^{1}$, Guilian Chen$^{1}$, Yuanzhu Gan$^{1}$, Lei Wang$^{1}$, Jian Pu$^{2}$
\thanks{$^{1}$Zongmu Technology}
\thanks{$^{2}$Fudan University}
}

\begin{document}

\maketitle
\thispagestyle{empty}
\pagestyle{empty}

\begin{abstract}
Multi-view radar-camera fused 3D object detection provides a farther detection range and more helpful features for autonomous driving, especially under adverse weather.
The current radar-camera fusion methods deliver kinds of designs to fuse radar information with camera data.
However, these fusion approaches usually adopt the straightforward concatenation operation between multi-modal features, which ignores the semantic alignment with radar features and sufficient correlations across modals.
In this paper, we present \textbf{MVFusion}, a novel \textbf{M}ulti-\textbf{V}iew radar-camera \textbf{F}usion method to achieve semantic-aligned radar features and enhance the cross-modal information interaction.
To achieve so, we inject the semantic alignment into the radar features via the semantic-aligned radar encoder (SARE) to produce image-guided radar features.
Then, we propose the radar-guided fusion transformer (RGFT) to fuse our radar and image features to strengthen the two modals' correlation from the global scope via the cross-attention mechanism.
Extensive experiments show that \textbf{MVFusion} achieves state-of-the-art performance (51.7\% NDS and 45.3\% mAP) on the nuScenes dataset. We shall release our code and trained networks upon publication.
\end{abstract}

\section{INTRODUCTION}

Significant progress of deep learning in recent years has achieved remarkable performance in the object detection task of autonomous driving~\cite{wang2021fcos3d,liu2022bevfusion}. 
To realize a more robust perception for driving assistance, the cooperation of multiple sensor modalities, such as camera, LiDAR, and millimeter wave
radar (radar), has attracted wide attentions in many research works~\cite{liang2018deep,2021RODNet,qian2021robust,li2022ezfusion}.

Each sensor modality has its own merits \cite{li2022ezfusion}.
Camera-based images provide high-resolution visual signals, which perform similar human visual perception, and even overcomes human limitation when equipped with multiple surround-view cameras.
LiDAR data supplements depth information compared to the single camera perception, but spends more cost \cite{nabati2021centerfusion,li2022ezfusion}.
However, the camera and LiDAR fail to offer robust perception under adverse weather, such as heavy rain, fog, and other low visibility conditions.
In contrast, radar reveals more stability to adverse conditions with lower cost \cite{nobis2019deepCRFNET}. 
In addition, radar provides a farther detection range and more helpful features, such as velocity, radar cross section (RCS), distance, etc.

Nevertheless, radar points perform sparse and noisy, lacking adequate vertical measurement, so researchers usually fuse the radar knowledge with LiDAR or camera data \cite{li2022ezfusion,lang2019pointpillars,dong2021radar}, where the radar-camera fusion reveals the mainstream strategy. 
For example, CenterFusion \cite{nabati2021centerfusion} achieves center-based radar and camera fusion via the specific two-stage design.
However, these fusion approaches usually adopt the straightforward concatenation operation between multi-modal features, which ignores the spatial connections and semantic alignment \cite{nabati2020radar,dong2021radar, nabati2021centerfusion}.

Moreover, the previous works \cite{kim2020grif, nabati2021centerfusion} focus on radar-camera fusion within monocular image content, there are few studies to survey the merging between multi-camera images and radar data.
Multi-view camera-based 3D object detection \cite{liu2022petr,huang2021bevdet} captures visual information across different camera views, and conducts more excellent performance than the monocular camera-based detection. 
Yet the multi-view radar-camera fusion remains under exploration, where radar data could assist and promote better surrounding perception.

\begin{figure}[tp!]
	\centering
	\includegraphics[width=0.95\linewidth]{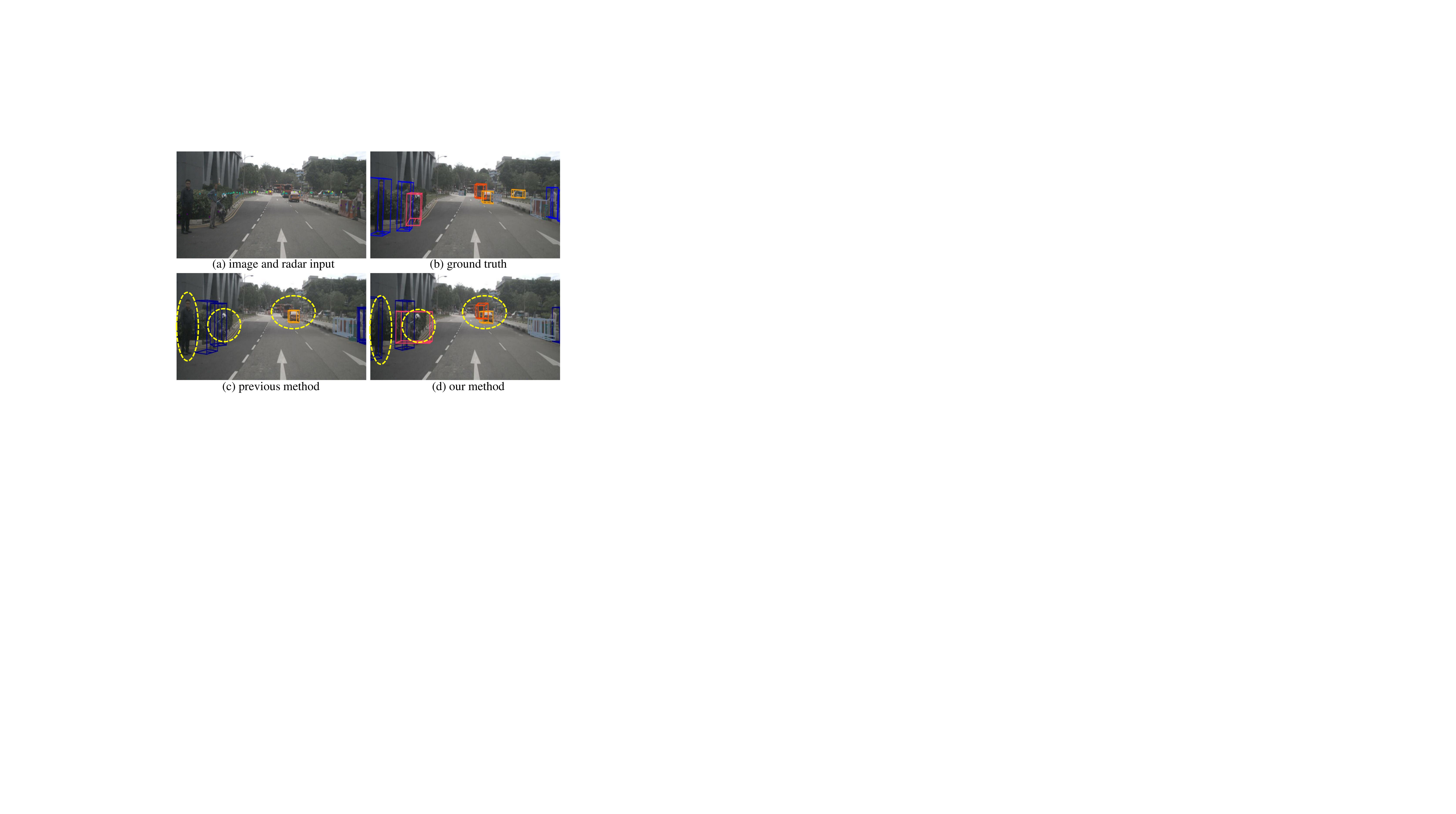}
	\vspace{-0.1in}
	\caption{Detection comparison between the camera-based approach \cite{liu2022petr} and our MVFusion.
(a) The image and radar input, where the color of radar points denotes the distance to the radar. 
(b) The 3D detection ground truth.
(c) The results of the camera-based approach \cite{liu2022petr}, which misses the detection of the distant car and the near pedestrian.
(d) Our method takes advantage of the semantic-aligned radar information to perform sufficient radar-camera fusion and successfully detects the missing car and pedestrian.
}
	\label{figue1}
	\vspace{-0.2in}
\end{figure}

In this paper, we present a novel multi-view radar-camera fusion method to achieve semantic-aligned radar features and enhance the cross-modal information interaction from the global scope.
As shown in Fig.\ref{figue1}, our method catches more helpful information via the semantic alignment radar features and radar-camera fusion, to receive more correct detection.

In detail, considering the lack of corresponding semantic information of radar points, firstly we propose the semantic-aligned radar encoder (SARE) to extract image-guided radar features, which adopts the semantic indicator to align radar inputs and select the image-guided radar transformer to form strong radar representations.

\begin{figure*}[t]
    \centering
    \includegraphics[width=0.8\textwidth]{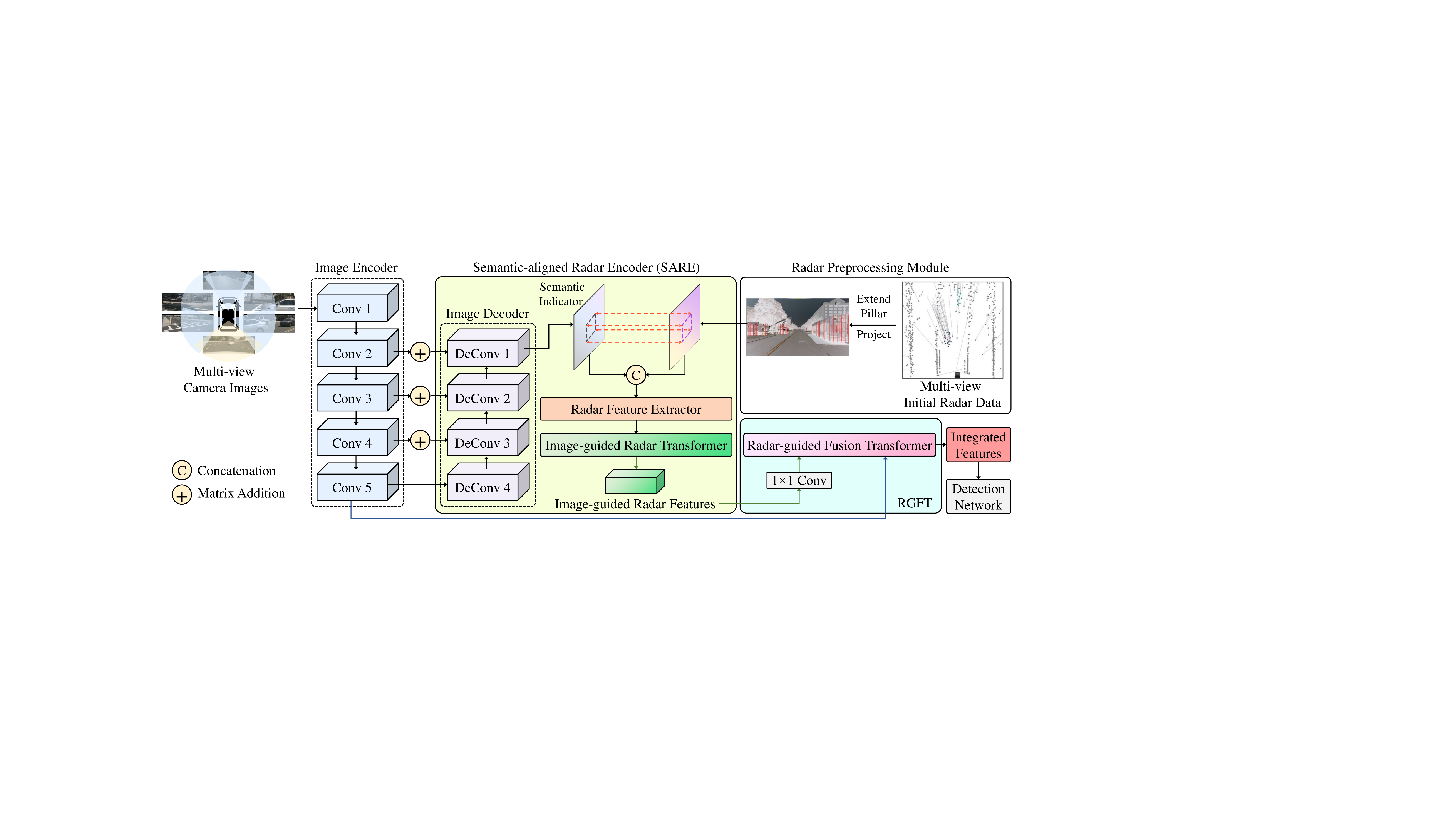}
    \vspace{-2mm}
    \caption{The overview of our proposed \textbf{MVFusion}, which mainly consists of five components: the radar preprocessing module, the image encoder, the semantic-aligned radar encoder (SARE), the radar-guided fusion transformer (RGFT) and the detection network.
    SARE injects the semantic alignment into the radar features, and RGFT fuses our radar and image features, aiming to sufficiently promote the two modal’s interaction from the global scope.
    The multi-view radar representation refers to \cite{schumann2021radarscenes}. 
    }
    \vspace{-0.2in}
    \label{fig:framework}
\end{figure*}

After radar feature extraction, we propose the radar-guided fusion transformer (RGFT) to integrate the enhanced radar features with high-level image features.
We utilize the cross-attention transformer to strengthen the cross-modal interaction from global scope, which performs beneficial for the subsequent 3D detection.

Our main contributions are as follows:

(1) We are the first to explore the radar-camera multi-view fusion for 3D object detection and propose our \textbf{M}ulti-\textbf{V}iew radar-camera \textbf{Fusion} approach: \textbf{MVFusion}, which utilizes visual semantics to obtain semantic-aligned radar features and applies robust fusion transformer to enhance the cross-modal information interaction.

(2) We propose the semantic-aligned radar encoder (\textbf{SARE}) to extract image-guided radar features, which adopts the semantic indicator to align radar data and image-guided radar transformer to produce strong radar representations.
Moreover, we present a radar-guided fusion transformer (\textbf{RGFT}) to integrate the enhanced radar features with high-level image features, to promote the two modal's sufficient correlation from the global scope via the cross-attention mechanism.

(3) Experiments show that \textbf{MVFusion} achieves state-of-the-art performance (51.7\% NDS and 45.3\% mAP) of the single-frame multi-view radar-camera fused 3D object detection on the standard nuScenes benchmark dataset.

\section{RELATED WORK}

\subsection{Single-modality 3D Object Detection}
Three-dimensional (3D) object detection reveals critical for various applications \cite{li2020rtm3d,li2021monocular,liu2021ground,reading2021categorical,wang2022probabilistic,lang2019pointpillars, zhou2018voxelnet}, such as autonomous driving, etc.
Previous works commonly utilize kinds of sensors (LiDAR, camera, and radar) to perform 3D object detection.
LiDAR-based methods \cite{lang2019pointpillars, zhou2018voxelnet, yan2018second,shi2019pointrcnn} have achieved the most precise 3D measurements but the most expensive price among the three sensors.
Then cameras achieve more affordable cost and provide abundant semantic information, though lacking depth information, which also attract extensive attention \cite{wang2021fcos3d, philion2020lift, wang2022detr3d, pan2020cross}, such as monocular 3D object detection \cite{liu2020smoke, reading2021categorical, wang2022probabilistic} and multi-view 3D object detection \cite{huang2021bevdet, wang2022detr3d, liu2022petr, li2022bevformer}.
Specifically, the multi-view 3D object detection integrates information across different camera views, for producing a more robust surrounding scene understanding.
BEVDet~\cite{huang2021bevdet} lifts the multi-view visual features to frustum features, then fuses and splats the frustum features into BEV (Bird's Eye View) space for detection.
DETR3D~\cite{wang2022detr3d} obtains 3D object queries from 2D features of multiple camera images and generates reference points from these queries.
PETR~\cite{liu2022petr} transforms the camera frustum space into 3D coordinates, to generate 3D position-aware features for the subsequent transformer decoder.
At last, radar provides important features like velocity, etc. in most weather conditions but achieves noisy position values, especially the height information \cite{nobis2019deepCRFNET}.
Researchers usually adopt the radar data together with other sensors' informantion to the multi-sensor fusion~\cite{nobis2019deepCRFNET, kim2020grif,long2021full,lin2020depth,dong2021radar,yang2020radarnet} detection.

\vspace{-0.1in}
\subsection{Fusion-based 3D Object Detection}
The 3D object detection based on multi-modal fusion collects single-modal advantages to promote the detection performance~\cite{li2022ezfusion,long2021full,lin2020depth,dong2021radar}.
Many works~\cite{chen2017multi,ku2018joint,vora2020pointpainting,li2022deepfusion,bai2022transfusion,wang2021pointaugmenting} focus on the fusion of camera and LiDAR data. 
For example, DeepFusion~\cite{li2022deepfusion} projects LiDAR points to the image plane and applies Learnable Align to capture the correlations with the cross-attention mechanism.
Pointaugmenting \cite{wang2021pointaugmenting} utilizes a BEVbased fusion approach with cross-LiDAR-camera data augmentation.
In addition, the immunity to extreme conditions and the characteristic cost-effective make radar considerable in developing autonomous driving with more robustness.
Recent methods~\cite{kim2020grif, nabati2020radar,dong2021radar, nabati2021centerfusion} have demonstrated the effectiveness based on the early, middle, and late fusion of radar and camera data.
The approach \cite{dong2021radar} focuses on object-level fusion via deep representation learning to explore the interaction and global reasoning of different sensor features.
CenterFusion~\cite{nabati2021centerfusion} detects objects' image centers to associate radar detections and generates radar features, then concatenates image and radar features for detection heads.
However, these fusion methods usually perform straightforward concatenation operations, ignoring the spatial connections and semantic alignment between radar and image features.
Besides, there are rare explorations for multi-view camera radar fusion.
So in this study, we aim to enhance the feature correlations and strengthen the spatial information interaction between radar and camera data from multiple views.

\vspace{-0.1in}
\section{METHOD}
\subsection{Overall Architecture}

We demonstrate the framework of our multi-view radar-camera fusion approach in Fig.~\ref{fig:framework}, which mainly consists of five components: the radar preprocessing module, the image encoder, the semantic-aligned radar encoder (SARE), the radar-guided fusion transformer (RGFT) and the detection network.
For the image inputs, given surround-view images from six views, we adopt the image encoder to extract the multi-scale image features, where we denote the image features of each stage as $F_{image}^s=\textrm{Conv}_s(I),\ s=1,2,3,4,5$.
For the radar points inputs from multi-view radars, we first employ the radar preprocessing module similar to \cite{nobis2019deepCRFNET} to obtain the radar representations with the same shape as images.
Then we apply our semantic-aligned radar encoder (SARE) to produce the image-guided enhanced radar features.
Afterward, we utilize the radar-guided fusion transformer (RGFT) to conduct the camera-radar fusion in terms of high-level features from the global scope, to create the integrated features.
Finally, the detection network chooses the cross-attention decoder and 3D detection head, to efficiently decodes the object queries and perform the final 3D predictions.

The radar preprocessing module exhibits essential, for the initial radar data reveals 3D points with additional properties, thus we need to transform them to a suitable formation to fuse the camera data (in this work, we project the radar points to the image space).
Moreover, the radar data performs sparse and noisy, so its direct utilization will affect the detection performance \cite{nobis2019deepCRFNET}.
Specifically, we obey two steps to pre-process the radar data.
Firstly, to avoid the noisy points disturbing, we filter the radar points with the 3D bounding boxes of ground truth, following \cite{nobis2019deepCRFNET}.
Besides, considering the incorrect or missing vertical measurement, we expand the points to the pillar with a three-meter height. 
Then we project the radar pillars to the corresponding image view with camera calibration, where we replace the back-view data with blank points, for the vacancy of radar at the back of the ego-vehicle.
Furthermore, we reserve another four significant characters to compose the five-channel radar maps ($I_{radar}$), where each point carries five properties: the position mask with zero or one (indicating the existence of radar points at the corresponding position of pixels); the distance; the radar cross section (RCS): a physical quantity of the echo intensity produced by the illumination of radar waves); the two radial velocities along X and Y axis.

\begin{figure}[t]
    \centering
    \includegraphics[width=0.9\linewidth]{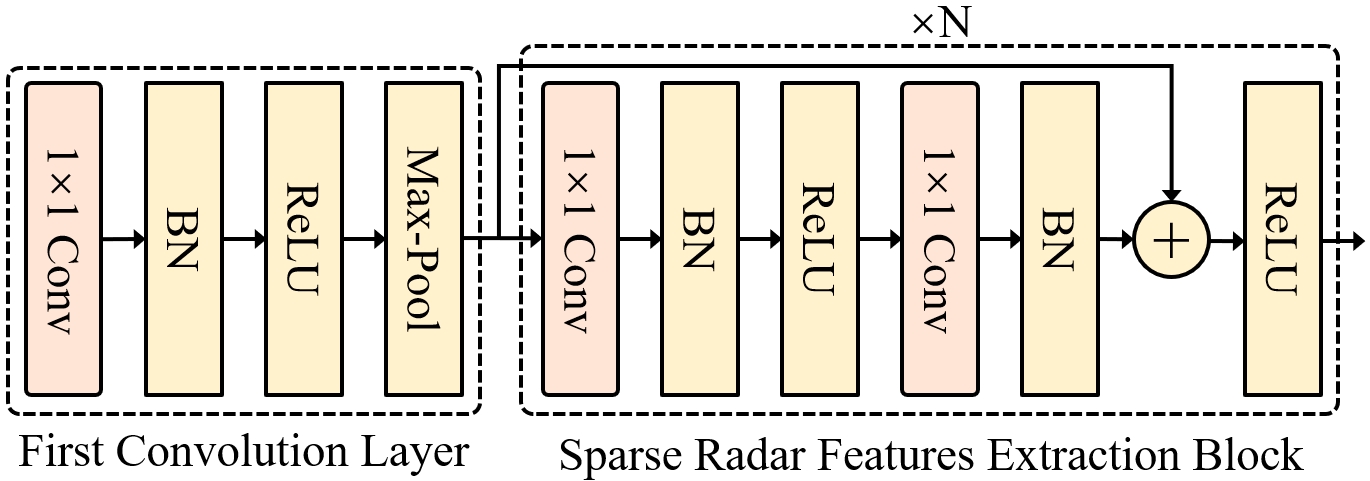}
    \vspace{-2mm}
    \caption{The schema of radar feature extractor (RFE), which consists of residual feature convolution blocks for the sparse radar features.}
    \label{fig:module3}
    \vspace{-0.2in}
\end{figure}

\subsection{Semantic-aligned Radar Encoder (SARE)}

Through the image encoder, we can obtain multi-scale image features with both low-level and high-level semantics.
Different from previous fusion methods \cite{kim2020grif, nabati2021centerfusion}, we firstly achieve semantic alignment between our five-channel radar maps after preprocessing and semantic image features before feeding into the radar feature extractor, as shown in Fig. \ref{fig:framework}.
Specifically, the semantic-aligned radar encoder contains three modules: the image decoder, the radar feature extractor and the image-guided radar transformer. 

To achieve semantic alignment, firstly we require a robust semantic indicator to align and filter the sparse radar inputs.
Here we produce our image indicator via all stages' visual features, where we upsample the high-level features through deconvolution blocks and complement them with original low-level features by skip-connections~\cite{ronneberger2015u}, to reach a more robust semantic indicator (SI). 
We denote the ${i}$-${th}$ output $D_i$ of the image decoder as:
\begin{equation}
    \textrm{D}_i=\left\{\begin{aligned}
        &\textrm{DeConv}(F_{image}^5),\quad i=4 \\
        &\textrm{DeConv}(\textrm{D}_{i+1})+F_{image}^{i+1},\quad i=1,2,3
    \end{aligned}
    \right.
\end{equation}
where $DeConv$ stands for the deconvolution block.
At last, we conduct 1$\times$1 convolution to reduce the channel to one and concatenate the image indicator to the radar inputs after prepossession.
We set the image indicator's channel as one, for we desire the indicator as the image foreground mask and utilize the foreground's semantics and relative location within the image space to achieve more robust alignment between semantic points and radar points. 
Concatenated under the high resolution, we succeed in accurately aligning radar points with the foreground semantics which are inferred according to the high-level visual features.


\begin{figure}[t]
    \centering
    \includegraphics[width=0.9\linewidth]{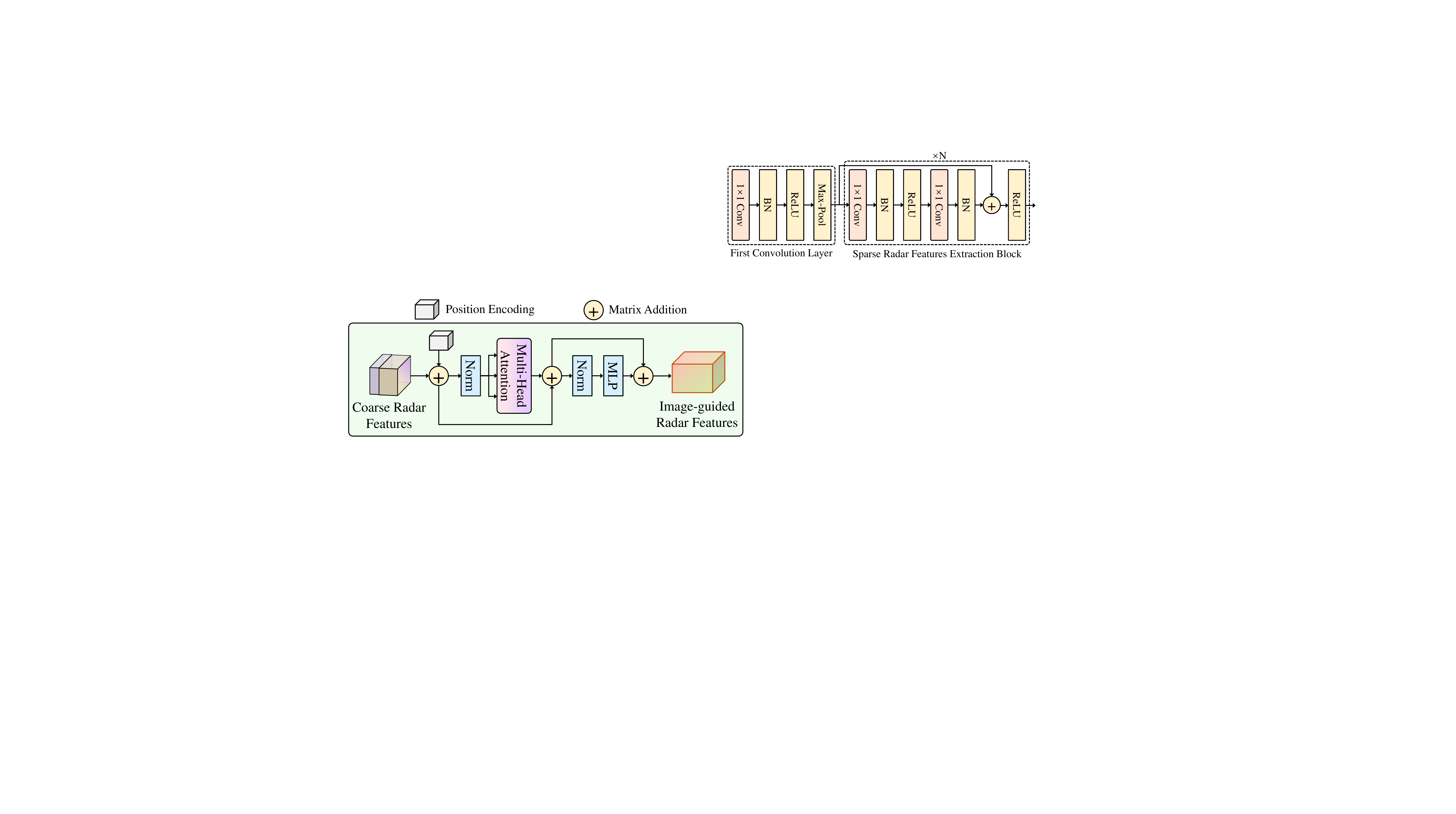}
    \vspace{-1mm}
    \caption{The overview of our image-guided radar transformer (IGRT). IGRT assigns learnable position encoding to the radar features to further enhance the spatial information through the multi-head self-attention mechanism.}
    \label{fig:module1}
    \vspace{-0.2in}
\end{figure}

After obtaining the semantic-aligned radar inputs, we put forward the radar feature extractor (RFE), as shown in Fig. \ref{fig:module3}. 
In the first convolution layer, we adopt the $1\times1$ convolution with stride of 2, and a max-pooling operation with kernel size of 2 to down-sample the features.
Furthermore, three sparse radar features extraction blocks with stride of 2 are employed to extract the final coarse radar features: $40\times100\times256$ ($H\times W \times C$).
We denote the corase radar features $F_{radar}=\textrm{RFE}(I_{indicator},I_{radar})$.

However, the universal convolution to extract radar features \cite{nabati2020radar,dong2021radar, nabati2021centerfusion, kim2020grif} performs the suboptimal performance, which ignores the implicit position and semantic relationships.
Thus we propose the image-guided radar transformer (IGRT) to model the long-range dependency and relationships for the coarse radar features.
In particular, firstly we flatten the features in $H$ and $W$ dimensions into the flattened sequences with the reshaping operation.
Then we introduce the multi-head self-attention mechanism to distribute attention on different positions simultaneously and enhance our radar features, as shown in Fig.~\ref{fig:module1}.
The self-attention achieves the overall and intelligent interaction to acquire sufficient correlations. 
Furthermore, we adopt the learnable position encoding \cite{VIT}, which reveals benefits for improving the spatial information learning of radar features.
We describe the self-attention procedure as follows:
\begin{align}
    Q,K,V=(F_{radar}+E_{pos})\textbf{W}_{Q,K,V}\\
    \textrm{IGRT-Attn} = \textrm{Softmax}(\frac{QK^T}{\sqrt{C/h}})V \\
    F_{radar}^{\prime} = \textrm{MLP}\big(\textrm{LN}(\textrm{IGRT-Attn})\big)
\end{align}
Q, K, and V denote the Query, Key, and Value.
$\textbf{W}_Q$, $\textbf{W}_K$ and $\textbf{W}_V$ are learnable parameters.
$E_{pos}$ stands for the learnable position encoding. 
$h$ denotes the number of heads for the multi-head self-attention. $\textrm{LN}$ means layer normalization and MLP means the multi-layer perceptron block.

\subsection{Radar-guided Fusion Transformer (RGFT)}

\begin{figure}[t]
    \centering
    \includegraphics[width=0.9\linewidth]{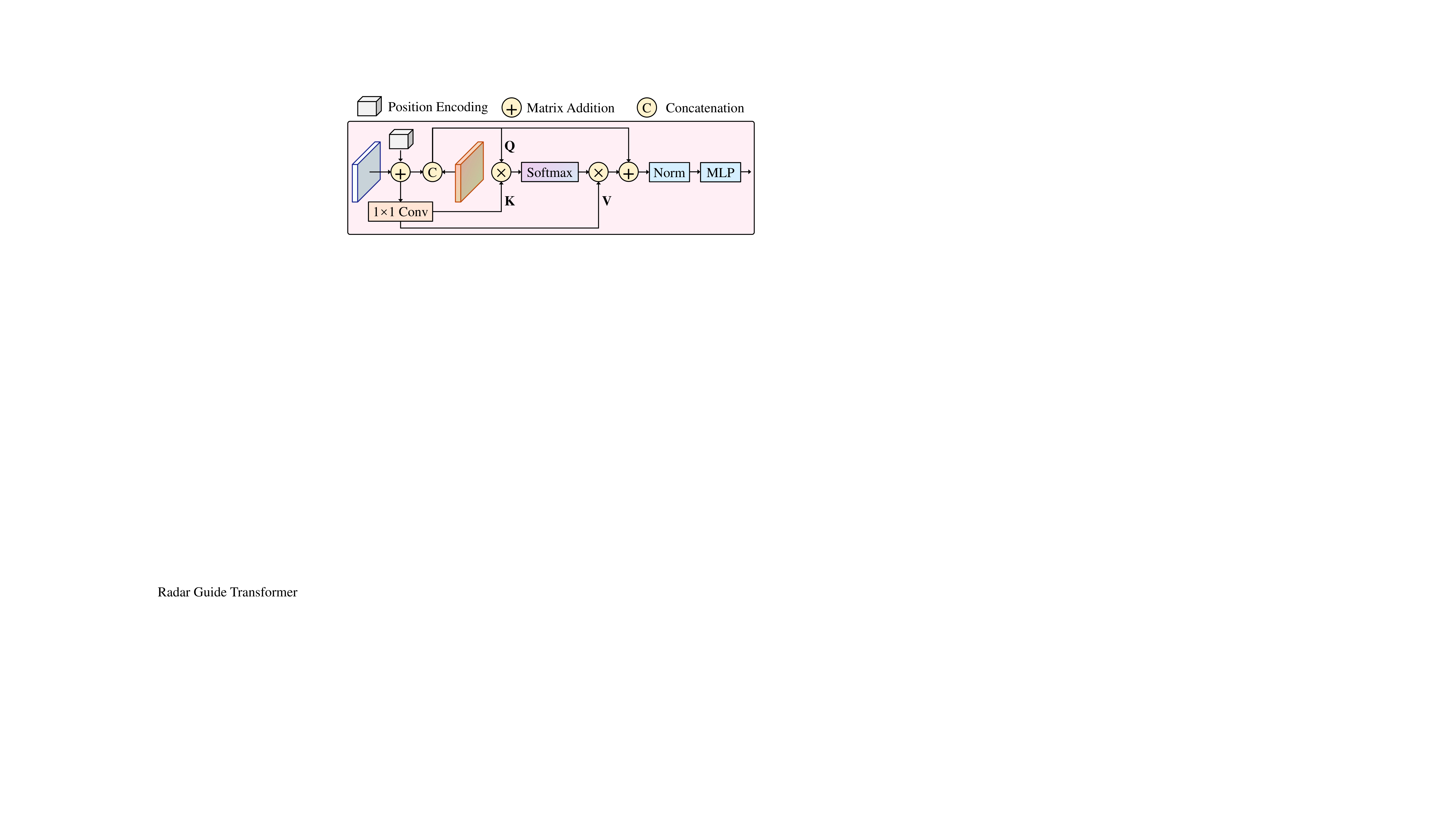}
    \vspace{-2mm}
    \caption{The overview of radar-guided fusion transformer (RGFT). 
    Fusing the high-level radar and image features, RGFT achieves adequate correlation under the cross-attention mechanism.}
    \label{fig:module2}
    \vspace{-0.2in}
\end{figure}

With the alignment and guidance of semantic information with image features, we attain enhanced and robust radar features.
To provide better fusion features for the detection network, we propose the radar-guided fusion transformer for camera-radar fusion in terms of high-level features from the global scope.
The earlier fusion approaches \cite{li2022ezfusion,lang2019pointpillars,dong2021radar} usually adopt the straightforward concatenation operation between multi-modal features, which ignores the sufficient and global spatial correlations across modals.
Notably, some transformer-based methods \cite{liu2022petr,li2022bevformer} succeed in managing the cross-modal problem via the cross-attention mechanism within the decoder. 
Thus here we introduce the cross-attention mechanism to fuse our radar and image features, aiming to strengthen the two modal's interaction from the global field, as shown in Fig. \ref{fig:module2}.


Different from the above radar encoder, for the fusion transformer, we remove the multi-head attention module to avoid the cost of large computation.
In addition, we utilize the 1$\times$1 convolution (Fig.~\ref{fig:framework}) to reduce the channel of high-level radar features before sending into the RGFT, and adopt another 1$\times$1 convolution (Fig.~\ref{fig:module2}) to expand the channel of $F_{image}^5$ to the fusion features' wanted channel.
\begin{align}
    &F_{radar}^{\prime\prime}=\textrm{conv}^a_{1\times1}(F_{radar}^{\prime})\\
    &F_{image}^{\prime\prime}=\textrm{conv}^b_{1\times1}(F_{image}^5+E'_{pos})
\end{align}
Furthermore, we adopt the concatenation of the radar and image features to create the cross-attention module's query, for only the relatively sparse radar features result in a convergence problem with exploding gradients (setting (c) in Table \ref{ablation_total2}).
On the other hand, we develop the keys and values from the image features.
Therefore, the cross-attention-related equation of our proposed RGFT can be defined as:
\vspace{-0.1in}
\begin{align}
    &Q=\textrm{Concat}(F_{radar}^{\prime\prime}, F_{image}^{\prime\prime}) \textbf{W}_{Q} \\
    &K=F_{image}^{\prime\prime}\textbf{W}_{K},\quad V=F_{image}^{\prime\prime}\textbf{W}_{V}\\
    &\textrm{RGFT-Attn} = \textrm{Softmax}(\frac{QK^T}{\sqrt{C}})V
\end{align}
where $E'_{pos}$ is the learnable position encoding for the image features, and $C$ denotes the channel number of $F_{image}^5$.
The final integrated features can be written as:
\begin{equation}
    F_{integrated} = \textrm{MLP}\big(\textrm{LN}(\textrm{RGFT-Attn})\big)
\end{equation}
With the radar-guided fusion transformer (RGFT) for adequate cross-modal correlations, we achieve robust integrated fusion features for the subsequent detection procedure.

\subsection{Detection Network}

We adopt the same multi-view decoder and 3D detection head as the previous work \cite{liu2022petr}, where object queries interact with the integrated fusion features through the multi-head attention and feed-forward network.
Finally, every object query predicts one 3D location and its category through the detection head, then we utilize the Hungarian algorithm \cite{kuhn1955hungarian} for label assignment between
ground truths and predictions.
We adopt the focal loss \cite{lin2017focal} for 3D classification and $L1$ loss for 3D regression:
\vspace{-0.1in}
\begin{equation}
    L_{3D} =\lambda_{cls}L_{cls} + \lambda_{reg}L_{reg}.
\end{equation}

\vspace{-0.1in}
\section{EXPERIMENTS}

\begin{table*}[t]
    \centering
    \caption{Comparison of single-frame state-of-the-art works on the nuScenes test set with different modals. 
    $\dag$ denotes adopting the DD3D \cite{park2021pseudo} pre-trained V2-99 \cite{lee2020centermask} backbone. 
    }
    \vspace{-0.1in}

    \begin{tabular}{l c l l l l l l l l l}
    \hline
      \textbf{Method}  & \textbf{Modal} & \textbf{NDS}$\uparrow$ & \textbf{mAP}$\uparrow$ & \textbf{mATE}$\downarrow$ & \textbf{mASE}$\downarrow$ & \textbf{mAOE}$\downarrow$ & \textbf{mAVE}$\downarrow$ & \textbf{mAAE}$\downarrow$ \\
    \hline
    InfoFocus \cite{wang2020infofocus} &L &0.395 &0.395 &0.363 &0.265 &1.132 &1.000 &0.395 \\
    
    PointPillars \cite{lang2019pointpillars}  &L &0.453 & 0.305 &0.517 & 0.290 &0.500 &0.316 &0.368 \\
    
    CenterFusion \cite{nabati2021centerfusion} & R\&C & 0.449 & 0.326 & 0.631 & 0.261 & 0.516 & 0.614 & 0.115 \\
    
    \hline
    
    CenterNet \cite{zhou2019objects}  &C & 0.400 & 0.338 & 0.658 & 0.255 & 0.629 & 1.629 & 0.142 \\
    
    FCOS3D \cite{wang2021fcos3d} & C & 0.428 & 0.358 & 0.690 & 0.249 & 0.452 & 1.434 & 0.124 \\
    
    PGD \cite{wang2022probabilistic} & C & 0.448 & 0.386 & 0.626 & 0.245 & 0.451 & 1.509 & 0.127 \\
    
    BEVDet \cite{huang2021bevdet} & C & 0.482 & 0.422 & 0.529 & 0.236 & 0.395 & 0.979 & 0.152 \\
    
    PETR \cite{liu2022petr}  & C  &0.483 &0.445 &0.627 &0.249 &0.449 &0.927 &0.141 \\

    DD3D$\dag$ \cite{park2021pseudo} & C & 0.477 & 0.418 & 0.572 & 0.249 & 0.368 & 1.014 & 0.124 \\
    
    DETR3D$\dag$ \cite{wang2022detr3d} & C & 0.479 & 0.386 & 0.626 & 0.245 & 0.394 & 0.845 & 0.133 \\
    
    BEVDet$\dag$ \cite{huang2021bevdet} & C & 0.488 & 0.424 & 0.524 & 0.242 & 0.373 & 0.950 & 0.148 \\
    
    PETR$\dag $ \cite{liu2022petr}   & C & 0.504  & 0.441 & 0.593 & 0.249 & 0.383 & 0.808 & 0.132 \\
     \hline
     \textbf{Ours}$\dag$ & R\&C & \textbf{0.517} &  \textbf{0.453}  & \textbf{0.569} & \textbf{0.246} & \textbf{0.379} & \textbf{0.781} & \textbf{0.128} \\
     \textcolor{blue}{Improvement}
     & \textcolor{blue}{vs. R\&C}
      & \textcolor{blue}{+6.8\%} &
      \textcolor{blue}{+12.7\%} &
      \textcolor{blue}{+6.2\%} &
      \textcolor{blue}{+1.5\%} &
      \textcolor{blue}{+13.7\%} &
      \textcolor{blue}{-16.7\%} &
      \textcolor{blue}{-1.3\%} 
      \\
      \textcolor{blue}{Improvement}
      & \textcolor{blue}{vs. C}
      & 
      \textcolor{blue}{+1.3\%} &
      \textcolor{blue}{+1.2\%} &
      \textcolor{blue}{+2.4\%} &
      \textcolor{blue}{+0.3\%} &
      \textcolor{blue}{+0.4\%} &
      \textcolor{blue}{+2.7\%} &
      \textcolor{blue}{+0.4\%} \\
      \hline
    \end{tabular}
    \label{table_test}
\end{table*}


\begin{table*}[t]
    \centering
    \caption{Comparison of single-frame state-of-the-art works on the nuScenes val set with different backbones and modals.  
    $\dag$ denotes adopting the DD3D \cite{park2021pseudo} pre-trained V2-99 \cite{lee2020centermask} backbone. 
    }
    \vspace{-0.1in}

    \begin{tabular}{l c c l l l l l l l l l}
    \hline
      \textbf{Method} &\textbf{Backbone} & \textbf{Modal} & \textbf{NDS}$\uparrow$ & \textbf{mAP}$\uparrow$ & \textbf{mATE}$\downarrow$ & \textbf{mASE}$\downarrow$ & \textbf{mAOE}$\downarrow$ & \textbf{mAVE}$\downarrow$ & \textbf{mAAE}$\downarrow$ \\
    \hline
    
    CenterNet \cite{zhou2019objects} &DLA & C &0.328 &0.306 &0.716 &0.264 &0.609 &1.426 &0.658  \\
    
    CenterFusion \cite{nabati2021centerfusion} &DLA & R\&C &0.453 &0.332 &0.649 &0.263 &0.535 &0.540 &0.142 \\
    
    \textbf{Ours} &DLA & R\&C & \textbf{0.471}
 & \textbf{0.419} & \textbf{0.664} & \textbf{0.258} & \textbf{0.416} & \textbf{0.789} & \textbf{0.193} \\
      & &\textcolor{blue}{Improvement}   &
      \textcolor{blue}{+1.8\%}&  
      \textcolor{blue}{+8.7\%} &
      \textcolor{blue}{-1.5\%} &
      \textcolor{blue}{+0.5\%} &
      \textcolor{blue}{+11.9\%} &
      \textcolor{blue}{-24.9\%} &
      \textcolor{blue}{-5.1\%} &
      \\
    
    \hline
    
    
    BEVDet \cite{huang2021bevdet} &Res-50 & C  &  0.381 &0.304  & 0.719 & 0.272 & 0.555 & 0.903 & 0.257 \\
    
    PETR \cite{liu2022petr} &Res-50 & C  &0.381 &0.313 &0.768 &0.278 &0.564 &0.923 &0.225   \\
      
     \textbf{Ours-R50} &Res-50 & R\&C & \textbf{0.402} & \textbf{0.330}  & \textbf{0.712} & \textbf{0.255} & \textbf{0.552} & \textbf{0.898} & \textbf{0.214} \\
    &
     &\textcolor{blue}{Improvement} & 
      \textcolor{blue}{+2.1\%} &
      \textcolor{blue}{+1.7\%} &
      \textcolor{blue}{+5.6\%} &
      \textcolor{blue}{+2.3\%} &
      \textcolor{blue}{+1.2\%} &
      \textcolor{blue}{+2.5\%} &
      \textcolor{blue}{+1.1\%} 
      \\
    \hline
    
    FCOS3D \cite{wang2021fcos3d} &Res-101 & C & 0.415 & 0.343 & 0.725 & 0.263 & 0.422 & 1.292 & 0.153  \\
    
    PGD \cite{wang2022probabilistic} &Res-101 & C & 0.428 & 0.369 & 0.683 & 0.260 & 0.439 & 1.268 & 0.185  \\
    
    DETR3D \cite{wang2022detr3d} &Res-101 & C & 0.434  & 0.349 & 0.716 & 0.268 & 0.379 & 0.842 & 0.200  \\

    PETR \cite{liu2022petr} &Res-101 & C  & 0.442 & 0.370 & 0.711 & 0.267 & 0.383 & 0.865 & 0.201  \\
    
    
    
     \textbf{Ours} &Res-101 & R\&C & \textbf{0.455} & \textbf{0.380} & \textbf{0.675} & \textbf{0.258} & \textbf{0.372} & \textbf{0.833} & \textbf{0.196} \\
    &
     &\textcolor{blue}{Improvement} & 
      \textcolor{blue}{+1.3\%} &
      \textcolor{blue}{+1.0\%} &
      \textcolor{blue}{+3.6\%} &
      \textcolor{blue}{+0.9\%} &
      \textcolor{blue}{+1.1\%} &
      \textcolor{blue}{+3.2\%} &
      \textcolor{blue}{+0.5\%} 
      \\
      \hline
      PETR\dag\cite{liu2022petr} &V2-99 & C & 0.455 & 0.403  & 0.736 & 0.271  & 0.432 & 0.825 & 0.204  \\
      \textbf{Ours}\dag&V2-99 & R\&C & \textbf{0.475}
 & \textbf{0.421} & \textbf{0.686} & \textbf{0.269} & \textbf{0.410} & \textbf{0.793} & \textbf{0.203} \\
    &
     &\textcolor{blue}{Improvement} & 
      \textcolor{blue}{+2.0\%} &
      \textcolor{blue}{+1.8\%} &
      \textcolor{blue}{+5.0\%} &
      \textcolor{blue}{+0.2\%} &
      \textcolor{blue}{+2.2\%} &
      \textcolor{blue}{+3.2\%} &
      \textcolor{blue}{+0.1\%} 
      \\
      \hline
    \end{tabular}
    \label{table_val}
    \vspace{-0.2in}
\end{table*}

\subsection{Datasets and Metrics}

We validate the effectiveness of our method on the public large-scale dataset nuScenes \cite{caesar2020nuscenes}, which contains a full 360-degree field of view provided by 6 cameras, 1 lidar and 5 radars. 
The dataset consists of 1000 driving scenes, with 700, 150 and 150 scenes for training, validation, and testing, respectively. 
We sample the corresponding sequences to frames at 2Hz and objects in each frame are annotated with 3D bounding boxes. 
We follow the official evaluation metrics of the dataset including nuScenes Detection Score (NDS) and mean Average Precision (mAP), along with mean Average Translation Error (mATE), mean Average Precision (mAP),  mean Average Translation Error (mATE), mean Average Scale Error (mASE), mean Average Orientation Error (mAOE), mean Average Velocity Error (mAVE), mean Average Attribute Error (mAAE).

\subsection{Implementation Details}
Our method applies ResNet \cite{ResNet} and VoVNetV2 \cite{lee2020centermask} as the feature extractor.
We adopt the same multi-view decoder and 3D detection head as the previous work \cite{liu2022petr}.
In addition, we also apply some basic experiment settings referred to \cite{liu2022petr}.
Specifically, we choose 64 points along the depth axis following the linear-increasing discretization (LID) \cite{tang2020center3d}, and set the region to $[- 61.2m, 61.2m]$ for the X and Y axis, and $[- 10m, 10m]$ for Z axis.
We set AdamW optimizer with the weight decay of 0.01 and initialize the learning rate with $2.0 \times 10^{-4}$. 
We select the $1600 \times 640$ resolution for nuScenes test set and different resolutions for nuScenes validation set.
We randomly rotate the inputs in the range $[-22.5 ^{\circ}, 22.5 ^{\circ}]$. 
We train all experiments for 24 epochs with a total batch size of 8 on 8 RTX A6000 GPUs. 

\subsection{Comparisons with State-of-the-art Methods}
Table \ref{table_test} shows the single-frame 3D object detection performance comparison on nuScenes \textbf{test} set.
Our method achieves the best performance on both NDS and mAP, compared with two Lidar-based methods InfoFocus \cite{wang2020infofocus}, PointPillars \cite{lang2019pointpillars}, the radar-camera fusion approach CenterFusion \cite{nabati2021centerfusion}, the camera-based monocular methods CenterNet \cite{zhou2019objects}, FCOS3D \cite{wang2021fcos3d}, PGD \cite{wang2022probabilistic}, DD3D \cite{park2021pseudo} and the camera-based multi-view methods DETR3D \cite{wang2022detr3d}, BEVDet \cite{huang2021bevdet} and PETR \cite{liu2022petr}. 
For the same radar-camera fusion, our approach exceeds the method CenterFusion \cite{nabati2021centerfusion} by 12.7\% in mAP and 6.8\% in NDS, respectively.
The CenterFusion \cite{nabati2021centerfusion} performs the two-stage fusion approach which proposes two heads to regress velocity and attributes, so their mAVE and mAAE achieve better results. 
Then, compared with the latest camera method PETR \cite{liu2022petr}, our approach achieves state-of-the-art performance: 1.2\% mAP and 1.3\% NDS.

\begin{figure*}[t]
\centering
    \includegraphics[width=0.80\textwidth]{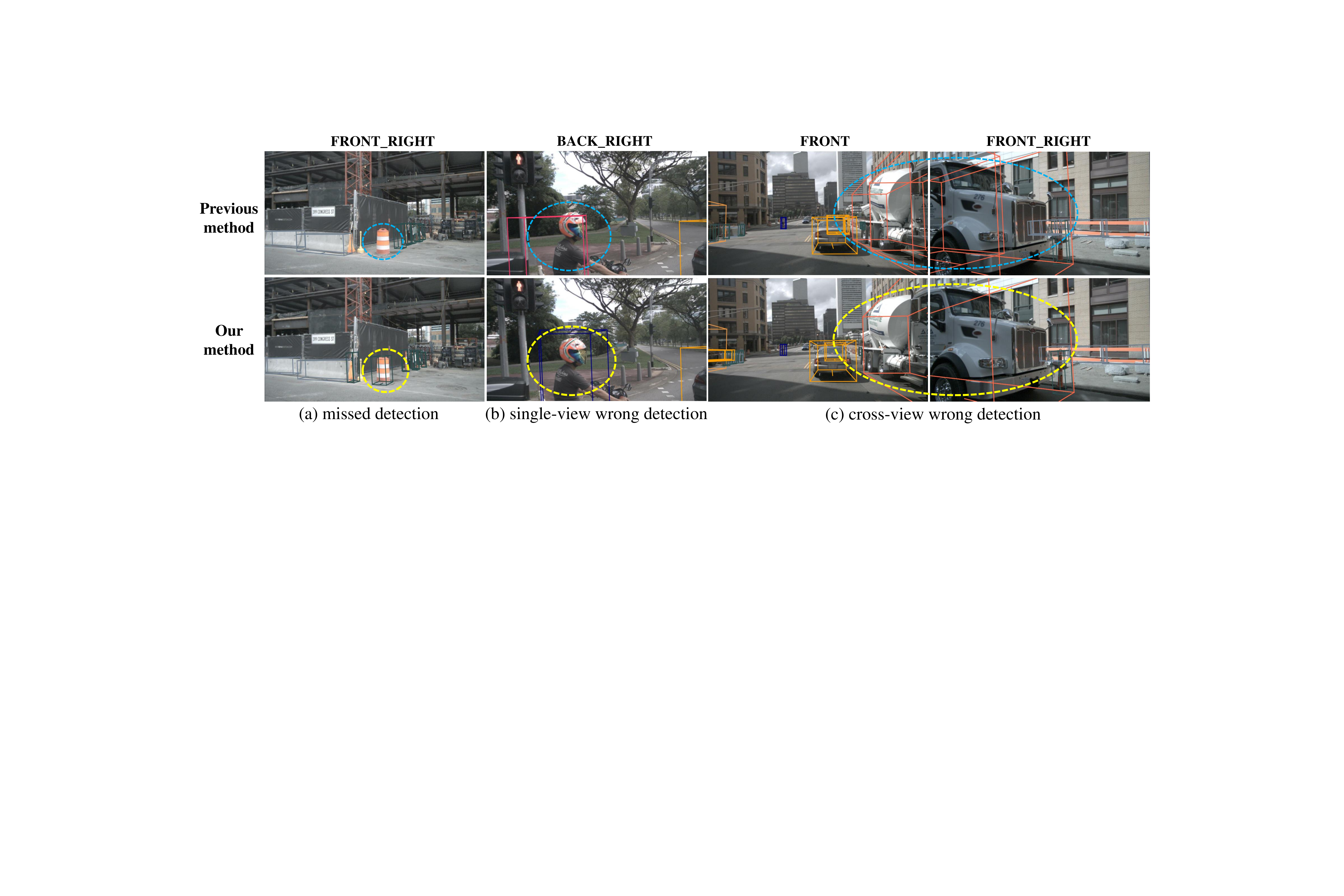}
    \vspace{-0.1in}
    \caption{
    The surround-view detection results comparison between our approach and the previous approach \cite{liu2022petr}.
    We mark the comparison areas with \textbf{yellow} circle for our approach and \textbf{blue} circle for \cite{liu2022petr}, respectively.
    Our approach achieves correct object detection for different views, where our sufficient radar-camera interaction between semantic-aligned radar features and visual features performs more helpful clues for 3D detection.
    }
    \label{fig_mask1}
    \vspace{-0.1in}
\end{figure*}

Furthermore, we compare the results with state-of-the-art methods on nuScenes \textbf{val} set, as shown in Table \ref{table_val}.
Following \cite{liu2022petr}, we compare our approach to previous methods with different backbones. 
It shows that our approach also achieves the best performance on both NDS and mAP.
Our approach surpasses the radar-camera fusion method CenterFusion \cite{nabati2021centerfusion} by 1.8 \% NDS and 8.7 \% mAP.
Our method also outperforms the camera-based methods by evenly about 3\% in mAP and about 2\% to 3\% in NDS.
The excellent performance on the nuScenes test and validation set strongly validates our approach's effectiveness.

\subsection{Ablation Study}
We perform ablation studies to validate the proposed modules, where we conduct all experiments using V2-99 \cite{lee2020centermask} on nuScenes val set \cite{caesar2020nuscenes}.

\subsubsection{Importance of proposed components.}
Table \ref{ablation_total} shows the ablation studies about our two modules: semantic-aligned radar encoder (SARE) and radar-guided fusion transformer (RGFT), which perform improvement on the detection performance.
Specifically, when inserting the SARE module (setting (b)), the semantic alignment and self-attention mechanism help achieve the semantic alignment for original sparse radar features to enhance the radar features' representations.
For RGFT (setting (c)), the fusion transformer brings global-scope adequate interaction, which reveals beneficial for detection.
Finally, our method achieves the highest accuracy when applying both two modules.

\begin{table}[t]
    \centering
    \caption{Ablation study on val set for proposed components.
    `SARE' denotes the semantic-aligned radar encoder, `RGFT' denotes the radar-guided fusion transformer.} 
    \vspace{-0.1in}
    \begin{tabular}{c|cc|cc} 
    \hline
       & \textbf{SARE} & \textbf{RGFT} & \textbf{NDS}$\uparrow$ & \textbf{mAP}$\uparrow$ \\
    \hline
     (a) &- &- & 0.457 & 0.406\\
     (b) &\checkmark  &- &0.467 & 0.415\\
     (c) &- &\checkmark  &0.465 & 0.414\\
     (d) & \checkmark &\checkmark & 0.475 & 0.421\\
    \hline
    \end{tabular}
\label{ablation_total}
\end{table}

\begin{table}[t]
    \centering
    \caption{Ablation study on val set for semantic-aligned radar encoder (SARE). 
    `SI' denotes the semantic indiciator. 
    `IGT' denotes the image-guided radar transformer.}
    \vspace{-0.1in}
    \begin{tabular}{c|cc|cc} 
    \hline
       & \textbf{SI} & \textbf{IGT} & \textbf{NDS}$\uparrow$ & \textbf{mAP}$\uparrow$ \\
    \hline
     (a) &- &- &0.465 & 0.414\\
     (b) &\checkmark &- &0.470 & 0.419\\
     (c) &-   &\checkmark  &0.468 & 0.417\\
     (d) & \checkmark &\checkmark  & 0.475 & 0.421\\
    \hline
    \end{tabular}
\label{ablation_total1}
\vspace{-0.2in}
\end{table}

\subsubsection{Ablation study for semantic-aligned radar encoder (SARE)}

We discuss the ablation effects within the proposed semantic-aligned radar encoder (SARE), as shown in Table \ref{ablation_total1}.
We remove all the SARE's settings as the baseline of Table \ref{ablation_total1} (setting (a)), which applies the radar inputs directly to the radar feature extractor with suboptimal grades.
The setting (b) validates that the semantic indicator succeeds in employing the foreground’s semantics and relative location to obtain the alignment.
And setting (c) reveals the image-guided radar transformer (IGRT), which promotes the long-range relationships for coarse radar features, performs a superior performance.
All settings of SARE (setting (d)) achieve the most enhancement. 

\subsubsection{Ablation study for radar-guided fusion transformer (RGFT)}
We investigate different settings within the radar-guided fusion transformer (RGFT), as shown in Table \ref{ablation_total2}.
We adopt the setting (a) without RGFT as the baseline of Table \ref{ablation_total2}.
Then setting (b) only utilizes the radar features as the RGFT's query, key and value, actually changing the RGFT to the self-attention, which improves limited performance, for we have conducted the self-attention for radar features in the above SARE.
So we bring the image features as key and value of RGFT (setting (c)), but the sparse radar features influence the RGFT's convergence and result in unstable training with exploding gradient.
Finally, we combine the radar and image features as queries and adopt the image features as key and value (setting (d)), to strengthen the two modal’s interaction from the global scope to reach the best performance.


\begin{table}[t]
    \centering
    \caption{Ablation study on val set for radar-guided fusion transformer (RGFT).
    `w' denotes `with' and `w/o' denotes `without'.
    `Q',`K',`V' denotes query, key, value. `img.' denotes image. `concat.` denotes `concatenation'.
    } 
    \vspace{-0.1in}
    \begin{tabular}{c|l|cc} 
    \hline
       & \textbf{Ablation} & \textbf{NDS}$\uparrow$ & \textbf{mAP}$\uparrow$ \\
    \hline
     (a) & w/o RGFT &0.467 & 0.415\\
     (b) & radar w/o img. &0.468 & 0.416\\
     (c) & radar as Q w img. as K,V  & N/A & N/A\\
     (d) & radar concat. img. as Q w img. as K,V   & 0.475 & 0.421\\
    \hline
    \end{tabular}
    \vspace{-0.2in}
\label{ablation_total2}
\end{table}

\subsection{Visualization}
We demonstrate the surround-view 3D detection visualization between our method and the previous approach \cite{liu2022petr} in Figure \ref{fig_mask1}.
Our approach achieves correct detection for the further, truncated, cross-view objects and releases the repeated prediction boxes.
The semantic-aligned radar features offer more helpful clues and supplement the single-modal visual features to conduct more robust 3D detection.

\section{CONCLUSION}
In this paper, we provide a novel \textbf{M}ulti-\textbf{V}iew radar-camera \textbf{Fusion} method \textbf{MVFusion} for 3D object detection, which achieve the semantic-aligned radar features and robust cross-modal information interaction.
Specifically, we propose the semantic-aligned radar encoder (SARE) to extract image-guided radar features.
After radar feature extraction, we propose the radar-guided fusion transformer (RGFT) to integrate the enhanced radar features with high-level image features.
Extensive experiments on the nuScenes dataset validate that our model achieves single-frame radar-camera fusion state-of-the-art performance.
In the future, we will aggregate the spatio-temporal information from multi-view cameras, to further promote the radar-camera fusion.


\begin{figure*}[t]
    \centering
    \includegraphics[width=0.90\textwidth]{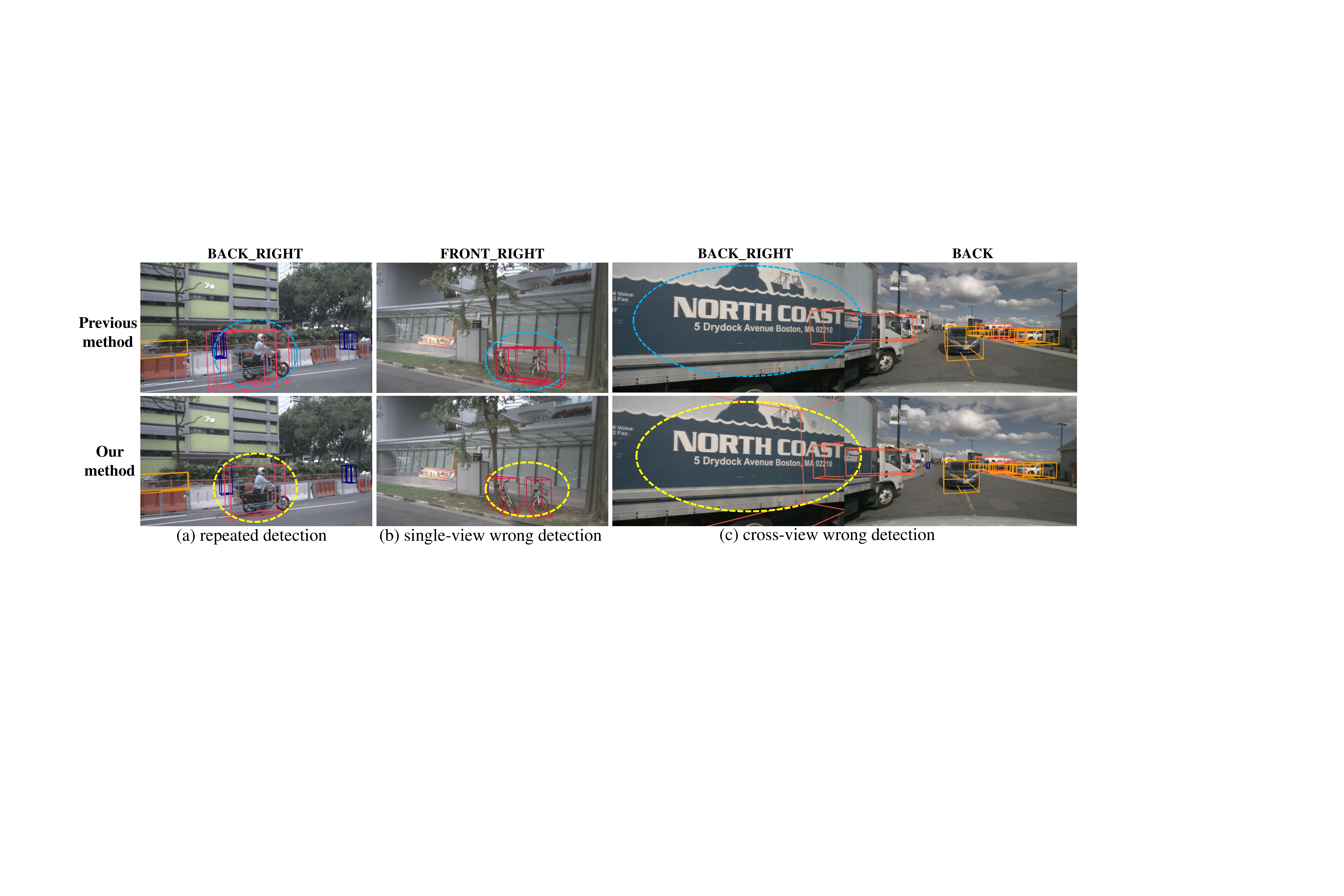}
    \caption{More surround-view detection results comparison between our approach and the previous approach \cite{liu2022petr}.
    We mark the comparison areas with \textbf{yellow} circle for our approach and \textbf{blue} circle for \cite{liu2022petr}, respectively.
    Our approach restraints the repeated detection, and release the single-view or cross-view wrong detection, where our adequate radar-camera interaction between semantic-aligned radar features and visual features performs the correct inference for 3D detection.
    }
    \label{fig:ref2}
\end{figure*}

\begin{figure*}[ht]
    \centering
    \includegraphics[width=0.98\textwidth]{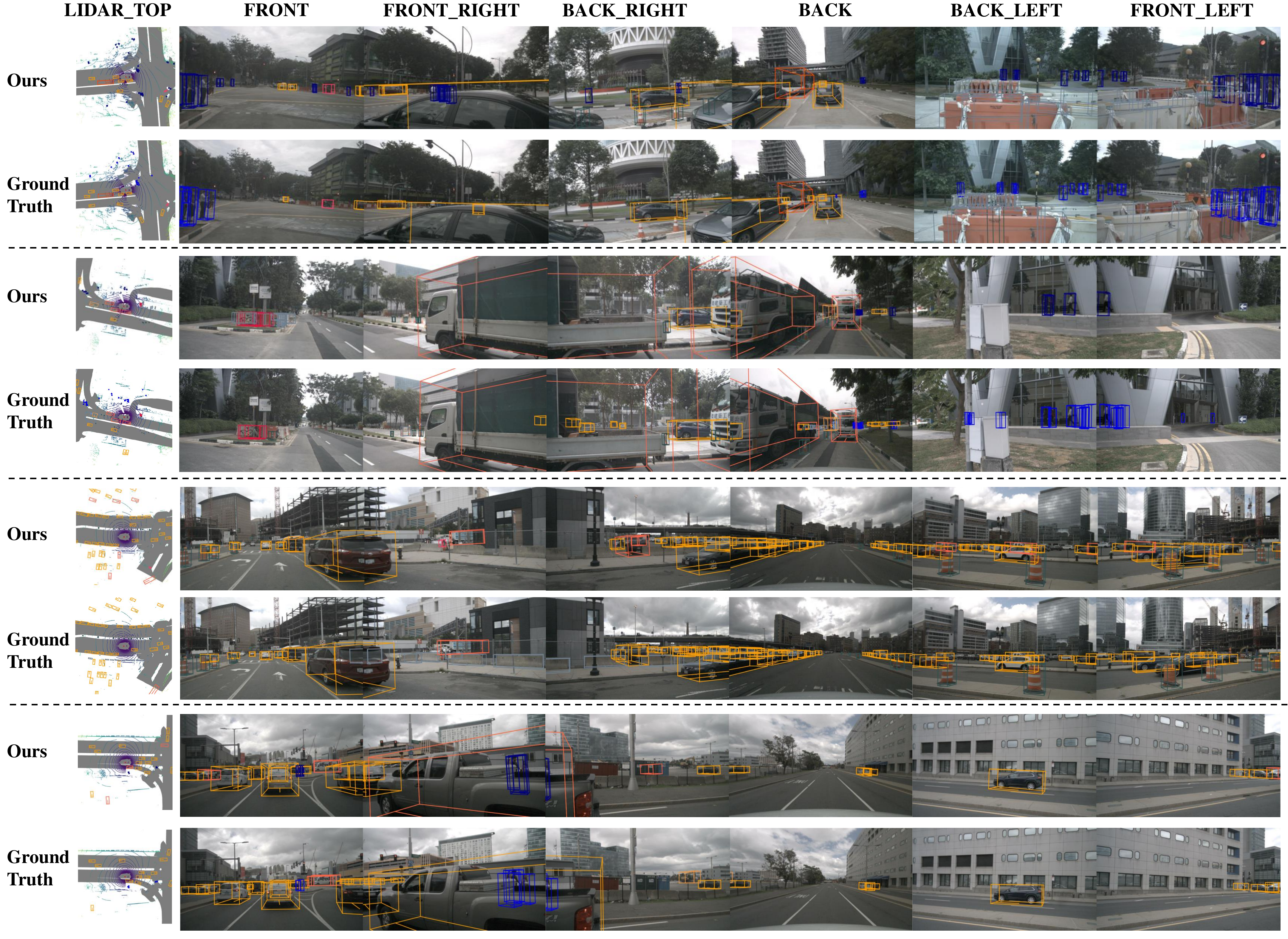}
    \caption{Qualitative comparison with ground truth. We compare 3D bounding boxes in image views and BEV views between our model's results and ground truth.
    The 3D bounding boxes are drawn with different colors to distinguish different classes.
    }
    \label{fig:ref3}
\end{figure*}

\section*{APPENDIX}


\subsection{Qualitative comparison with previous method}

We provide more qualitative comparisons between our method and the previous approach \cite{liu2022petr}, as shown in Figure \ref{fig:ref2}, where our method achieves success on further, truncated, partially occluded, cross-view objects and restraints the repeated prediction boxes with the semantic-aligned radar features and robust cross-modal information interaction.

\subsection{Qualitative comparison with ground truth}

We provide more qualitative analyses of our model compared with ground truth. As shown in Figure \ref{fig:ref3}, our 3D detection results are close to the ground truth. 









\clearpage
\bibliographystyle{IEEEtran}

\end{document}